\pdfoutput=1
\documentclass{article}

\usepackage[numbers]{natbib}
\usepackage[preprint]{neurips_2024}

\usepackage[utf8]{inputenc} 
\usepackage[T1]{fontenc}    
\usepackage[hidelinks]{hyperref}
\usepackage{url}            
\usepackage{booktabs}       
\usepackage{amsfonts}       
\usepackage{nicefrac}       
\usepackage{microtype}      
\usepackage[dvipsnames]{xcolor}         
\usepackage{pgfplots}
\usepackage{longtable}
\usepgfplotslibrary{external}
\usepackage{multirow}
\usepackage{array}     
\usepackage{amsmath}
\usepackage{float}
\usepackage{parcolumns}
\usepackage{caption}   
\usepackage{tikz}
\usepackage{geometry}
\usepackage{graphicx} 
\usepackage{enumitem}
\title{Beyond Captioning: Task-Specific Prompting for Improved VLM Performance in Mathematical Reasoning}

\author{%
 Ayush Singh, Mansi Gupta, Shivank Garg, Abhinav Kumar, Vansh Agrawal \\
  Vision and Language Group\\
  Indian Institute of Technology, Roorkee\\
  \texttt{\{ayush\_s@mt,m\_gupta@ma,shivank\_g@mfs,abhinav\_k@ma,vansh\_a@ph\}.iitr.ac.in} \\
}

\begin{document}

\maketitle

\begin{abstract}
Vision-Language Models (VLMs) have transformed tasks requiring visual and reasoning abilities, such as image retrieval and Visual Question Answering (VQA). Despite their success, VLMs face significant challenges with tasks involving geometric reasoning, algebraic problem-solving, and counting. These limitations stem from difficulties in effectively integrating multiple modalities and accurately interpreting geometry-related tasks \cite{zhang2024vision}. Various works claim that introducing a captioning pipeline before VQA tasks enhances performance \cite{ozdemir2024enhancing}. We incorporated this pipeline for tasks involving geometry, algebra, and counting and found that captioning results are not generalizable specifically with larger VLMs primarily trained on downstream QnA tasks showing random performance on math-related challenges. However, we present a promising alternative: task-based prompting, enriching the prompt with task-specific guidance. This approach shows promise and proves more effective than direct captioning methods for math-heavy problems.
\end{abstract}


\section{Introduction}
\label{Intro}

With the rise of Large Language Models, which demonstrate the ability to understand and generate text for tasks beyond their explicit training, Vision-Language Models have extended these capabilities to multimodal tasks involving images and text \cite{zhang2024visionlanguagemodelsvisiontasks}. These models excel in tasks like Visual Question Answering (VQA), image captioning, and object segmentation \cite{qiao2024prism}. However, recent studies reveal that VLMs struggle with simple, low-level visual tasks that humans easily solve, highlighting a need to enhance their visual reasoning and understanding \cite{rahmanzadehgervi2024vision}. 

Some works try to improve VLMs' reasoning abilities by fine-tuning the VLMs \cite{roberts2024smart}, \cite{yang2024mathglm},\cite{deng2024enhancing}. Additionally, some research on VLMs focuses on improving their question-answering abilities through a two-step process: captioning followed by question-answering. This method takes advantage of VLMs' pre-training in text generation, as many tasks require generating text descriptions. The main challenge lies in the model's ability to effectively combine and interpret multimodal information, understanding visual and textual inputs while capturing their interactions. One area where VLMs consistently underperform is counting \cite{paiss2023teaching}, primarily due to the scarcity of training data that accurately labels object counts, especially as the number of objects increases. While captioning has improved performance in some tasks \cite{ozdemir2024enhancing}, we hypothesize that these improvements are not generalizable and depend on various factors, which we aim to explore through our experiments. Further, captioning fails to capture all the attributes of the image, which is especially crucial in mathematical tasks.

To address these limitations, we introduce prompting techniques designed to enhance the models' reasoning capabilities. we specifically constructed prompts based solely on the question, excluding any direct information about the answer. These approach-based prompts were tested in both direct QnA tasks and as guides for captioning, with the expectation of improving performance. Additionally, we assessed robustness by using adversarial prompts, which provide incorrect problem-solving strategies but relevant to the problem, and random prompts to introduce irrelevant text, evaluating the models' responses and robustness to these perturbations.

\subsection{Background and Related work}
Several studies have examined the reasoning and comprehension abilities of Vision-Language Models in various tasks requiring spatial understanding and reasoning capabilities. These studies have demonstrated that multimodal language models rely less on visual information and perform better when they are given adequate textual cues.\cite{zhang2024can}, \cite{wang2024picture}. While methods like few-shot prompting \cite{NEURIPS2020_1457c0d6} have been shown to improve the performance of VLMs \cite{chen2023measuring}, these models continue to struggle with mathematical tasks, particularly counting, leading some to describe them as ``blind'' to numbers \cite{rahmanzadehgervi2024vision}. Recent research suggests that much of the reasoning performed by VLMs may stem more from the phrasing of the questions rather than the images themselves. This is evident in tasks that heavily rely on visual information, such as counting nested squares or identifying line intersections, where VLMs consistently underperform. Datasets like Math Vision \cite{wang2024measuring} and Count Bench\cite{paiss2023teaching} have been developed specifically to test these visual reasoning abilities. 

To enhance Visual Questioning (VQA) performance, various techniques have been proposed, including the use of question-driven image captions, which are subsequently fed into language models. These approaches have demonstrated potential to enhance outcomes in specific tasks, such as direct image-based QnA. \cite{ozdemir2024enhancing}. However, whether such captioning-based techniques can reliably enhance VLM performance on math-related tasks has been explored in our work.
\section{Method}
\label{Method}
We assessed the Vision-Language Models on a range of geometry-related tasks. We take tasks from four datasets containing different questions to ensure our tests' robustness and generalization. These datasets contain a variety of tasks related to geometry, counting, algebra, and mathematical reasoning. We use a diverse set of VLMs in our experiments to assess the generalizability of our approach. One closed-source large model was taken, Gemini-1.5-Flash, and three open-sourced smaller models, LLaVa, Florence-2, and Phi 3.5 Vision Instruct, were chosen. Such a range of models was chosen to ensure variation in size, from smaller ones with fewer parameters to larger, more complex models. They were tested across eight distinct tasks, divided into two main categories: 

\begin{enumerate}
    \item \textbf{Question-Answering (QnA):} Using a classical zero-shot approach, each model was directly queried with questions related to images from the datasets.
    \item \textbf{Captioning:} We generated captions for the image using the base model. After generating the captions, we fed them back into the LLM, and QnA was performed on the generated caption using the LLM.
\end{enumerate}

\begin{figure}[h]
\centering
\includegraphics[width=1.0\textwidth, height=0.21\textheight]{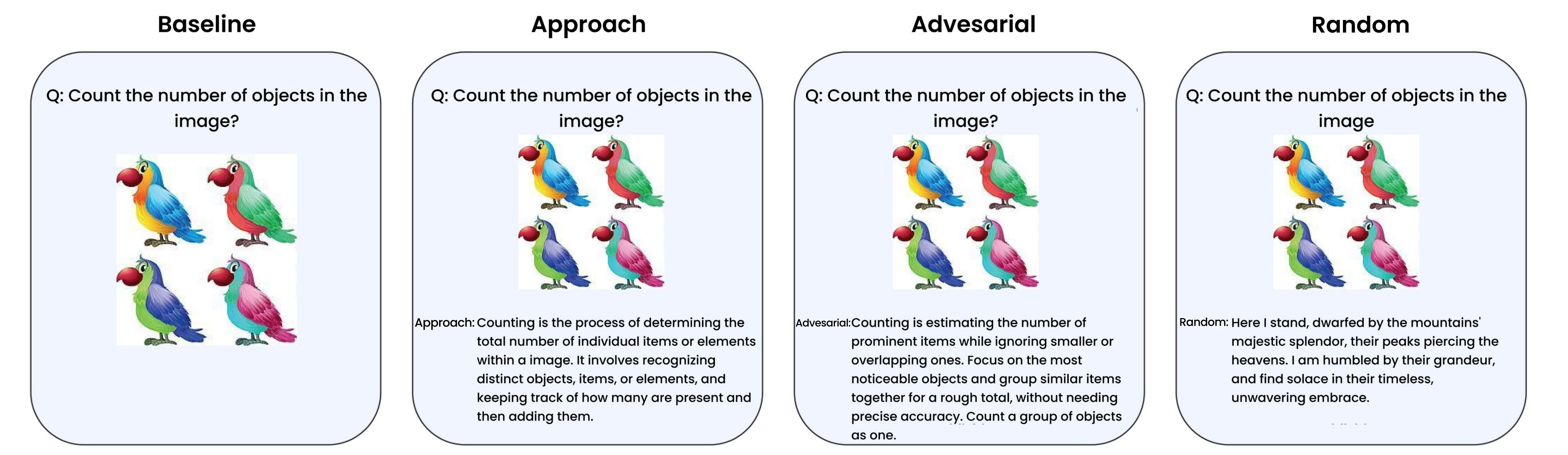} 
\caption{Example of our QnA Approach}
\label{Approach}
\end{figure}

We further tested the impact of incorporating additional information and context into the prompts. Specifically, we provided explicit guidance to solve the problem, which was generated using an LLM(Gemini).
This method aimed to determine how effectively the models could leverage explicit procedural guidance to enhance their performance on QnA. Additionally, we tested two other variants, random prompts and adversarial prompts(Figure: \ref{Approach}). Exact details regarding which are mentioned in Appendix: \ref{prompt}.

In our captioning experiments, first, we generated image captions by inputting task-specific keywords derived from the Llama 3.1-Instruct model \cite{dubey2024llama}. These keywords were extracted by prompting the model to produce concise, 1-2-word summaries that encapsulate the essence of each question. After generating the captions, we fed them back to an LLM and asked the corresponding questions for each task. Similar to Direct QnA, we try approach-based captioning where, along with the image and keywords, the approach was passed to the model to generate the image caption, which is further used to extract the final answer.

\begin{figure}[h]
\centering
\includegraphics[width=0.80\textwidth, height=0.2\textheight]{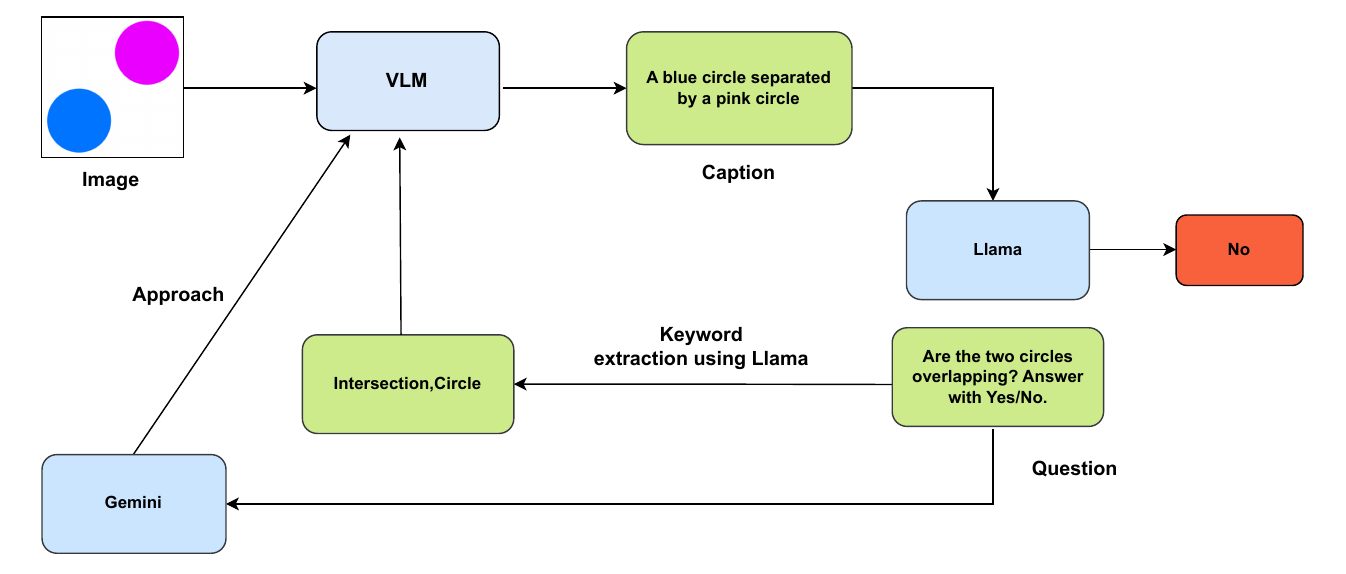} 
\caption{Example of our Caption Based Approach}
\label{Approach}
\end{figure}

\section{Experiments and Results}
\label{expt and result}

\begin{figure}[h]
\centering
\begin{tikzpicture}
\begin{axis}[
        width=0.5\textwidth,
        height=5cm,
        ymin=0,
        ymax=50,
        xticklabels={Gemini, LLaVa, Florence, Phi-3.5, cot},
        x tick label style={rotate=45,anchor=east,
                /pgf/number format/1000 sep=},
        ylabel=Accuracy,
        enlargelimits=0.05,
        legend style={at={(0.5,-0.4)},
        anchor=north,legend columns=-1},
        ybar interval=0.4,
        legend image code/.code={
            \draw[#1] (0cm,-0.1cm) rectangle (0.2cm,0.1cm);
        },
]
\addplot
        coordinates {(1, 42.73611111) (2, 9.70277777)
                 (3, 7.122222222) (4, 28.08888889) (5,0)};
\addplot
        coordinates {(1,44.321) (2,12.538)
                 (3,7.493) (4,32.488) (5,0)};
\legend{Base, Approach}
\end{axis}
\end{tikzpicture}
\hfill
\begin{tikzpicture}
\begin{axis}[
        width=0.5\textwidth,
        height=5cm,
        ymin=0,
        ymax=50,
        xticklabels={Gemini, LLaVa, Florence, Phi-3.5, cot},
        x tick label style={rotate=45,anchor=east,
                /pgf/number format/1000 sep=},
        ylabel=Accuracy,
        enlargelimits=0.05,
        legend style={at={(0.5,-0.4)},
        anchor=north,legend columns=-1},
        ybar interval=0.4,
        legend image code/.code={
            \draw[#1] (0cm,-0.1cm) rectangle (0.2cm,0.1cm);
        },
]
\addplot
        coordinates {(1, 42.73611111) (2, 9.70277777)
                 (3, 7.122222222) (4, 28.08888889) (5,0)};
\addplot
        coordinates {(1,44.321) (2,12.538)
                 (3,7.493) (4,32.488) (5,0)};
\legend{Base, Approach}
\end{axis}
\end{tikzpicture} 
\caption{Left: Comparison of captioned methods,Right: Comparison of base methods, }
\label{base_app}
\end{figure}
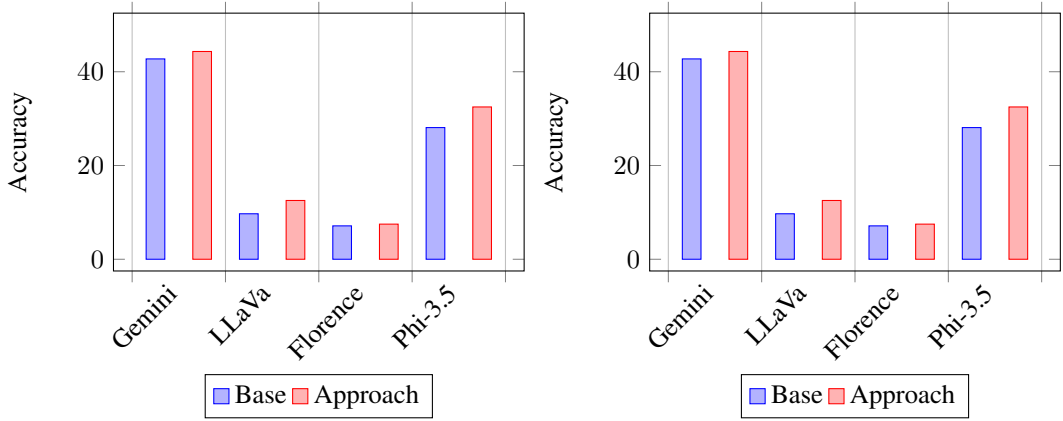

We chose diverse models and techniques to prove and test our hypothesis. We chose four datasets, Geo170k, CountBench, Blind, and MathVision, containing various tasks related to geometry, reasoning, algebra, and counting. Also, we split the MathVision dataset into three subparts: mainly vision-based, geometry-based, and mathematics-based (exact details of the datasets in Appendix: \ref{Dataset}). The approach-based prompting improves overall results for both Direct VQA and Caption-based QA (Figure: \ref{base_app}). Further, we observe a drop in performance when prompted with the adversarial approach and an overall increase in performance compared to baseline when prompted with the random approach(Table: \ref{Prompt table}). We also observe that the performance of the models varies across different datasets(Table: \ref{Data table}). Models perform best on the CountBench dataset, which focuses on counting tasks, while they perform poorly on the MathVision Dataset due to the complexity of the tasks. Further, on the MathVision dataset, we observe better performance on visual-based tasks as compared to mathematics-related tasks. Also, performance on geometry related tasks 
is observed to be relatively poor. (Appendix: \ref{Dataset})


 \begin{table*}[htbp]
    \centering
    \small 
    \setlength{\tabcolsep}{4pt} 
    \begin{tabular}{lcccccccc}
    \toprule
    \textbf{Dataset} & \textbf{Base} & \textbf{Approach} & \textbf{Random} & \textbf{Adv.} & \textbf{Caption} & \textbf{Caption} & \textbf{Caption} & \textbf{Caption} \\
    & & & & & & \textbf{Approach} & \textbf{Adv.} & \textbf{Random} \\
    \midrule
    \textbf{Gemini-1.5-Flash} & 42.73 & \textbf{44.32} & 41.30 & 43.86 & 35.18 & \textbf{42.53} & 37.70 & 38.62 \\
    \textbf{LLaVa} & 09.70 & \textbf{12.54} & 09.89 & 10.61 & 24.06 & \textbf{25.54} & 23.15 & 25.07 \\
    \textbf{Florence-2} & 07.12 & \textbf{07.49} & 04.89 & 02.89 & 14.03 & \textbf{16.86} & 14.56 & 16.35 \\
    \textbf{Phi-3.5-Vision} & 28.09 & \textbf{32.49} & 30.58 & 28.92 & 31.44 & \textbf{34.76} & 29.27 & 28.71 \\
    \bottomrule
    \end{tabular}
    \caption{Model-wise comparison of accuracy of the different approaches}
    \label{Prompt table}
\end{table*}

 \begin{table*}[htbp]
    \centering
    \small 
    \setlength{\tabcolsep}{4pt} 
    \begin{tabular}{lcccccccc}
    \toprule
    \textbf{Dataset} & \textbf{Base} & \textbf{Approach} & \textbf{Random} & \textbf{Adv.} & \textbf{Caption} & \textbf{Caption} & \textbf{Caption} & \textbf{Caption} \\
    & & & & & & \textbf{Approach} & \textbf{Adv.} & \textbf{Random} \\
    \midrule
    \textbf{Math-vision} & 10.13 & \textbf{14.06} & 11.92 & 13.31 & 19.28 & \textbf{22.31} & 20.43 & 19.61 \\
    \textbf{CountBench} & 32.58 & \textbf{33.33} & 31.67 & 34.50 & 27.16 & \textbf{34.16} & 28.00 & 31.00 \\
    \textbf{Geo} & 16.22 & \textbf{20.46} & 19.11 &  14.67 & 24.00 & \textbf{28.56} & 26.00 & 23.00 \\
    \textbf{Blind} & 31.52 & \textbf{32.12} & 27.08 & 25.98 & 31.94 & \textbf{32.81} & 28.60 & 31.90 \\
    \bottomrule
    \end{tabular}
    \caption{Dataset-wise comparison of accuracy of the different approaches}
    \label{Data table}
\end{table*}
\section{Conclusion}
\label{conclusion}

The results of our study align with our initial hypothesis: VLMs exhibit significant limitations when it comes to mathematical tasks, particularly those involving numbers and counting. While captioning using task-specific keywords is a first step to improve performance in some contexts, our findings suggest that its effectiveness is inconsistent, varying greatly depending on the dataset, task complexity, and model size. Larger models, often pre-trained on QnA tasks, inherently perform better in QnA-related tasks, a trend not observed in smaller models highlighting the influence of pre-training, as observed in Gemini.(Further refer Appendix:\ref{Experiments})

Our experiments demonstrate the potential for improving VLMs through techniques that enhance their reasoning capabilities and consistently improve performance. Additionally, in our assessment for testing the models' robustness and generalization capabilities using adversarial and random approach techniques, we observed a drop in performance, supporting our claim that VLMs incorporate additional information into their reasoning. Meanwhile, the performance stability when random prompts were provided underscores the robustness of these models’ reasoning abilities(Appendix: \ref{random_app}). In conclusion, improving the reasoning capabilities of VLMs presents a promising path forward, especially given their inherent blindness in perceiving numbers and handling mathematical tasks. By leveraging structured prompts constructed solely by leveraging information from the question, and approaches that guide reasoning, we can mitigate some of these weaknesses and move closer to enhancing their performance in more complex problem-solving scenarios.
\section{Limitations and Future Work}
\label{Future}
Testing the reasoning capabilities of multimodal models is a broad area of research, and we propose ways to improve the generalizability of these models in our research. Due to resource constraints, we couldn't experiment with many mainstream and large-scale models, especially state-of-the-art Vision-Language Models (VLMs). With more computational resources and funding, future work could focus on scalability and robustness across a wider range of model architectures such as Claude Sonnet\cite{claude}, GPT 4\cite{achiam2023gpt}, etc.

Additionally, advanced methods, such as sophisticated prompt engineering techniques or incorporating domain-specific knowledge—could enhance the models' ability to generate more accurate and contextually relevant captions. Moreover, replacing the QA model with interpretable alternatives could offer greater transparency and insights into the decision-making process, thus shedding light on the model's reasoning and performance in a more understandable way.

\bibliography{bibliography}
\bibliographystyle{unsrt}

\newpage
\appendix
\section{Prompting}
\label{prompt}
To enrich the prompts used in our experiments, we employed Gemini 1.5 Flash. The question was presented to the model without an accompanying image. The model was tasked with generating responses under three different conditions:
\begin{enumerate}
    \item \textbf{Approach-Based:} The model was asked to provide an approach for solving the question.
    \item \textbf{Adversarial:} The model was prompted to generate a misleading or incorrect approach to solving the question. Although inaccurate, the response needed to be plausible.
    \item \textbf{Random:} The model was asked to generate a random string.
\end{enumerate}
In the Captioning-based experiments, we utilized the LLaMA 3.1 Instruct 8B\footnote{https://huggingface.co/meta-llama/Llama-3.1-8B-Instruct} model to generate keywords based solely on the question. These keywords provided initial guidance for the captioning task. For the zero-shot captioning, we instructed the respective model being tested to generate a caption for the image using the keywords provided. Similarly, for the rest of the captioning tasks, we asked the QnA base model to generate image captions, using both keywords and additional hints generated in the same manner as described above (Approach-Based, Adversarial, Random), following which the answer was generated by passing the caption to another LLM(Llama 3.1 Instruct). \newline
\textbf{Note:} For Florence-2 captions were generated using <DETAILED CAPTION>, and the approach was passed onto the detailed caption. For other models, the approach was passed during the caption generation stage. Additionally, the Florence-2 direct checkpoint was unable to perform QnA-related tasks, so we used Florence-2 DocQnA for QnA-related tasks.

\section{Experiment Details}
\label{Experiments}
Certain experimental details worth mentioning for the respective models used are:
\begin{itemize}
 \item \textbf{ LLaVa:} For LLaVa, we used the GrokAPI \footnote{https://developers.x.ai/python-sdk/grok/} to access the model. 
 \item \textbf{Gemini-1.5-Flash:} For Gemini, we used Google AI studio\footnote{https://aistudio.google.com/} to access the model.
  \item \textbf{Florence-2:\footnote{https://huggingface.co/microsoft/Florence-2-large}\footnote{https://huggingface.co/HuggingFaceM4/Florence-2-DocVQA}} For Florence-2, we used the open-sourced model available on huggingface .
  \item \textbf{Phi 3.5 Vision Instruct:\footnote{https://huggingface.co/microsoft/Phi-3.5-vision-instruct}} we used the open-sourced model available on huggingface.
 \end{itemize}
\section{Datasets}
\label{Dataset}
The following datasets were used for our experiments:
 \begin{itemize}
 \item \textbf{Math Vision:\footnote{https://huggingface.co/datasets/MathLLMs/MathVision}} The Math Vision dataset
 is a curated collection of 3,040 high-quality mathematical problems with visual contexts from real math competitions. For our experiments, we broadly divided the dataset into three categories:
 \begin{itemize}
 \item \textbf{Visual Based:} This was originally split into Area, Angles, and Length-related tasks.
  \item \textbf{Geometry Based:} This was originally split into categories: Analytical Geometry, Combinatorial Geometry, Transformation Geometry, Descriptive Geometry and Solid Geometry.
  \item \textbf{General Mathematics:} This was originally split into categories: Graph Theory, Logic, Algebra, Combinatorics, Statistics, and Arithmetic.
 \end{itemize}
  
 \item \textbf{Blind:\footnote{https://huggingface.co/datasets/XAI/vlmsareblind}}The Blind dataset consisted of images and question-answer pairs about visual tasks. We used a subset of 150 images per task. The tasks include counting the number of intersections of 2 circles or lines, checking if 2 lines are intersecting, counting the number of rows and columns in a grid, finding the number of overlapping circles in an image, and finding the number of paths between 2 points in a subway connection image.
 
  \item \textbf{Countbench:\footnote{https://huggingface.co/datasets/nielsr/countbench}}The CountBench dataset contained a total of 540 images containing between two and ten instances of a particular object, where their corresponding captions reflect this number. This dataset is a benchmark dataset for counting related tasks.
  \item \textbf{Geo170k:\footnote{https://huggingface.co/datasets/Luckyjhg/Geo170K}}The Geo dataset contained more than 170K geometric image-caption and question-answer pairs. We used a subset of 500 images to conduct our experiments.
 \end{itemize}

\section{The Random Approach}
\label{random_app}
When using the random approach, we observed that performance often surpassed the baseline, with significant improvements in some cases. We hypothesize that the model tends to disregard random information, but in doing so, it becomes more cautious and focused on providing a correct response. This contrasts with the baseline, where such behavior is less apparent. While these findings are promising, there is still potential for further research to understand and refine this approach fully. 
\section{Task Wise Keywords}
\label{prompt-table}

\begin{table}[h]
\centering
\begin{tabular}{lll}
\hline
\textbf{Dataset name} & \textbf{Task in dataset} & \textbf{Keywords for task} \\
\hline
 & & \\
\multirow{14}{*}{MathVision} & Analytic Geometry & analytic geometry \\
 & Algebra & algebra, mathematics, logic \\
 & Transformation Geo & transformation, geometry \\
 & Statistics & statistics, graph \\
 & Angle & metric geometry, angles, mathematics, logic \\
 & Combinatorics & combinatorics, logic \\
 & Descriptive Geo & descriptive geometry, mathematics \\
 & Logic & logics, reasonings \\
 & Length & lengths, geometry \\
 & Arithmetic & arithmetic, logics, mathematics \\
 & Area & area, geometry \\
 & Combinatorial Geo & combinatorial, geometry \\
 & Solid Geometry & solids, geometry \\
 & Graph Theory & logics, connections, graphs \\
 & & \\
\hline
 & & \\
Countbench & Counting objects & counting, objects \\
 & & \\

\hline
 & & \\
GEO170K & Geometry problems & geometry, mathematics \\
 & & \\
\hline
 & & \\
\multirow{5}{*}{BLIND} & Line intersection & count number, intersections \\
 & Two line intersection & lines, intersecting \\
 & Interior pentagon & count, number, pentagons \\
 & Subway & subway lines, count, paths \\
 & Rows/columns & count, rows, columns \\
 & Two circles & circles, touching \\
 & & \\
\hline
\end{tabular}
\caption{Dataset Information}
\label{tab:dataset_info}
\end{table}
\section{Results Table}
\label{final-table}

\begin{table}[htbp]
\centering
\resizebox{\textwidth}{!}{
\tiny
\begin{tabular}{lcccccccc}
\hline
\textbf{Task} & \textbf{0-Shot} & \textbf{Adv} & \textbf{Random} & \textbf{Caption} & \textbf{App-Capt} & \textbf{Adv-Cap} & \textbf{Rand-Cap} & \textbf{Caption} \\
\hline
\multicolumn{9}{c}{\textbf{MathVision}} \\
\hline
Angles & 36 & 36 & 24 & 38 & 36 & 30 & 34 & 32 \\
Area & 27 & 28 & 24 & 28 & 32 & 34 & 28 & 22 \\
Length & 30 & 32 & 24 & 36 & 30 & 26 & 32 & 36 \\
Descriptive Geo & 34 & 26 & 28 & 22 & 14 & 20 & 24 & 20 \\
Analytic Geo & 16 & 18 & 22 & 14 & 20 & 22 & 20 & 12 \\
Combinatorial Geo & 22 & 26 & 14 & 26 & 20 & 24 & 10 & 6 \\
Transformation Geo & 18 & 24 & 20 & 22 & 20 & 14 & 28 & 22 \\
Solid Geo & 20 & 32 & 20 & 24 & 16 & 28 & 16 & 14 \\
Graph Theory & 28 & 24 & 22 & 26 & 24 & 26 & 22 & 18 \\
Arithmetic & 26 & 26 & 28 & 20 & 22 & 36 & 26 & 18 \\
Logic & 32 & 32 & 36 & 14 & 18 & 36 & 22 & 30 \\
Combinatorics & 20 & 20 & 12 & 22 & 22 & 28 & 14 & 12 \\
Algebra & 26 & 22 & 30 & 22 & 18 & 32 & 28 & 10 \\
Statistics & 26 & 24 & 24 & 24 & 22 & 24 & 22 & 12 \\
\hline
\textbf{Dataset Average} & 26.44 & 27.29 & 23.38 & 25.64 & 23.89 & 27.31 & 24.42 & 20.49 \\
\hline
\multicolumn{9}{c}{\textbf{GEO170K}} \\
\hline
Geometry problems & 30 & 33.33 & 32 & 30 & 30 & 36 & 34 & 28 \\
\hline
\multicolumn{9}{c}{\textbf{CountBench}} \\
\hline
Counting objects & 62 & 68 & 64.67 & 68 & 43 & 62.66 & 52 & 64 \\
\hline
\multicolumn{9}{c}{\textbf{Blind}} \\
\hline
Line Intersection & 50 & 58 & 44 & 59 & 43 & 43 & 35 & 38 \\
Two line intersection & 72 & 70 & 54 & 72 & 73 & 74 & 70 & 71 \\
pentagon & 53 & 42 & 24 & 51 & 45 & 41 & 43.34 & 42 \\
Subway & 23 & 23 & 26 & 16 & 0 & 0 & 0 & 0 \\
Rows/columns & 28 & 32 & 31 & 31 & 19 & 24 & 11 & 18 \\
Two circles & 89 & 67 & 92 & 82 & 83 & 83 & 83 & 83 \\
\hline
\textbf{Dataset Average} & 52.50 & 48.67 & 45.16 & 51.84 & 43.87 & 44.16 & 40.39 & 42.00 \\
\hline
\end{tabular}
}
\caption{Gemini Model Performance Across Various Tasks}
\label{tab:gemini-performance}
\end{table}

\begin{table}[htbp]
\centering
\resizebox{\textwidth}{!}{
\tiny
\begin{tabular}{lcccccccc}
\hline
\textbf{Task} & \textbf{0-Shot} & \textbf{Adv} & \textbf{Random} & \textbf{Caption} & \textbf{App-Capt} & \textbf{Adv-Cap} & \textbf{Rand-Cap} & \textbf{Caption} \\
\hline
\multicolumn{9}{c}{\textbf{MathVision}} \\
\hline
Angles & 6 & 14 & 8 & 14 & 38 & 34 & 28 & 26 \\
Area & 2 & 4 & 2 & 8 & 24 & 26 & 26 & 28 \\
Length & 6 & 14 & 14 & 14 & 20 & 26 & 28 & 26 \\
Descriptive Geo & 20 & 12 & 16 & 16 & 20 & 22 & 24 & 26 \\
Analytic Geo & 4 & 12 & 8 & 10 & 22 & 12 & 26 & 20 \\
Combinatorial Geo & 12 & 14 & 12 & 14 & 16 & 12 & 8 & 12 \\
Transformation Geo & 18 & 16 & 14 & 14 & 24 & 24 & 28 & 42 \\
Solid Geo & 4 & 10 & 6 & 6 & 22 & 26 & 20 & 16 \\
Graph Theory & 12 & 6 & 6 & 4 & 18 & 24 & 28 & 12 \\
Arithmetic & 2 & 14 & 10 & 6 & 16 & 22 & 14 & 22 \\
Logic & 10 & 16 & 8 & 8 & 16 & 24 & 18 & 26 \\
Combinatorics & 4 & 2 & 8 & 4 & 16 & 14 & 20 & 20 \\
Algebra & 0 & 4 & 4 & 4 & 26 & 28 & 16 & 20 \\
Statistics & 6 & 12 & 6 & 12 & 20 & 12 & 14 & 20 \\
\hline
\textbf{Dataset Average} & 7.31 & 10.82 & 8.73 & 10.11 & 22.27 & 22.84 & 22.29 & 23.29 \\
\hline
\multicolumn{9}{c}{\textbf{GEO170K}} \\
\hline
Geometry problems & 5 & 8 & 6.67 & 6 & 33.33 & 36 & 32.66 & 36 \\
\hline
\multicolumn{9}{c}{\textbf{CountBench}} \\
\hline
Counting objects & 12 & 14 & 10 & 12 & 14.66 & 16 & 14 & 15 \\
\hline
\multicolumn{9}{c}{\textbf{Blind}} \\
\hline
Line Intersection & 4 & 1 & 3 & 8 & 6 & 5 & 6 & 8 \\
Two line intersection & 32 & 42 & 30 & 23 & 64 & 68 & 55 & 60 \\
pentagon & 2 & 3 & 2 & 5 & 11 & 10 & 9 & 3 \\
Subway & 6 & 8 & 7 & 8 & 28 & 25 & 26 & 26 \\
Rows/columns & 5 & 9 & 7 & 9 & 6 & 10 & 6 & 8 \\
Two circles & 38 & 41 & 36 & 33 & 41 & 46 & 40 & 51 \\
\hline
\textbf{Dataset Average} & 14.50 & 17.33 & 14.16 & 14.33 & 26 & 27.33 & 23.67 & 26 \\
\hline
\end{tabular}
}
\caption{LLAVA Model Performance Across Various Tasks}
\label{tab:llava-performance}
\end{table}

\begin{table}[htbp]
\centering
\resizebox{\textwidth}{!}{
\tiny
\begin{tabular}{lcccccccc}
\hline
\textbf{Task} & \textbf{0-Shot} & \textbf{Adv} & \textbf{Random} & \textbf{Caption} & \textbf{App-Capt} & \textbf{Adv-Cap} & \textbf{Rand-Cap} & \textbf{Caption} \\
\hline
\multicolumn{9}{c}{\textbf{MathVision}} \\
\hline
Angles & 5 & 2 & 0 & 3 & 22 & 24 & 17 & 26 \\
Area & 4 & 3 & 0 & 4 & 22 & 26 & 24 & 28 \\
Length & 6 & 4 & 0 & 2 & 16 & 26 & 25 & 26 \\
Descriptive Geo & 3 & 3 & 1 & 4 & 16 & 24 & 0 & 27 \\
Analytic Geo & 2 & 5 & 2 & 6 & 17 & 16 & 20 & 21 \\
Combinatorial Geo & 3 & 2 & 1 & 3 & 19 & 15 & 8 & 10 \\
Transformation Geo & 1 & 2 & 1 & 3 & 20 & 23 & 18 & 22 \\
Solid Geo & 4 & 3 & 0 & 2 & 19 & 21 & 19 & 18 \\
Graph Theory & 1 & 2 & 0 & 2 & 17 & 23 & 28 & 12 \\
Arithmetic & 2 & 3 & 1 & 3 & 15 & 23 & 18 & 21 \\
Logic & 2 & 2 & 1 & 3 & 15 & 24 & 20 & 25 \\
Combinatorics & 2 & 4 & 1 & 2 & 15 & 15 & 17 & 19 \\
Algebra & 2 & 3 & 0 & 2 & 24 & 25 & 18 & 19 \\
Statistics & 1 & 1 & 1 & 2 & 18 & 16 & 16 & 19 \\
\hline
\textbf{Dataset Average} & 3.09 & 2.83 & 0.56 & 2.98 & 18.51 & 22.04 & 19.25 & 21.81 \\
\hline
\multicolumn{9}{c}{\textbf{GEO170K}} \\
\hline
Geometry problems & 0 & 1.34 & 0 & 0 & 4 & 9 & 6 & 5 \\
\hline
\multicolumn{9}{c}{\textbf{CountBench}} \\
\hline
Counting objects & 3 & 4 & 2 & 4 & 6 & 8 & 8 & 9 \\
\hline
\multicolumn{9}{c}{\textbf{Blind}} \\
\hline
Line Intersection & 15 & 10 & 31 & 0 & 37 & 38 & 30 & 39 \\
Two line intersection & 74 & 76 & 31 & 0 & 69 & 70 & 69 & 73 \\
pentagon & 0 & 0 & 0 & 0 & 0 & 0 & 0 & 0 \\
Rows/columns & 0 & 0 & 0 & 0 & 0 & 0 & 0 & 0 \\
Two circles & 23 & 23 & 23 & 23 & 32 & 34 & 26 & 36 \\
\hline
\textbf{Dataset Average} & 22.4 & 21.8 & 17 & 4.6 & 27.6 & 28.4 & 25 & 29.6 \\
\hline
\end{tabular}
}
\caption{Florance-2 Model Performance Across Various Tasks}
\label{tab:florence-2-performance}
\end{table}

\begin{table}[htbp]
\centering
\resizebox{\textwidth}{!}{
\tiny
\begin{tabular}{lcccccccc}
\hline
\textbf{Task} & \textbf{0-Shot} & \textbf{Adv} & \textbf{Random} & \textbf{Caption} & \textbf{App-Capt} & \textbf{Adv-Cap} & \textbf{Rand-Cap} & \textbf{Caption} \\
\hline
\multicolumn{9}{c}{\textbf{MathVision}} \\
\hline
Angles & 0 & 12 & 18 & 14 & 10 & 20 & 18 & 14 \\
Area & 2 & 18 & 16 & 14 & 14 & 20 & 18 & 16 \\
Length & 6 & 22 & 14 & 14 & 14 & 24 & 18 & 10 \\
Descriptive Geo & 6 & 16 & 20 & 16 & 16 & 22 & 22 & 16 \\
Analytic Geo & 2 & 10 & 14 & 12 & 14 & 14 & 26 & 14 \\
Combinatorial Geo & 4 & 18 & 10 & 18 & 12 & 12 & 8 & 12 \\
Transformation Geo & 4 & 18 & 24 & 16 & 14 & 10 & 8 & 12 \\
Solid Geo & 6 & 14 & 12 & 14 & 4 & 6 & 4 & 4 \\
Graph Theory & 0 & 16 & 14 & 14 & 10 & 16 & 14 & 8 \\
Arithmetic & 8 & 20 & 16 & 18 & 20 & 16 & 18 & 20 \\
Logic & 6 & 12 & 14 & 14 & 10 & 22 & 18 & 14 \\
Combinatorics & 6 & 14 & 12 & 14 & 14 & 14 & 14 & 12 \\
Algebra & 2 & 12 & 18 & 14 & 4 & 10 & 12 & 8 \\
Statistics & 2 & 6 & 4 & 12 & 18 & 24 & 18 & 20 \\
\textbf{Dataset Average} & 3.68 & 15.28 & 15 & 14.51 & 12.44 & 17.04 & 15.75 & 12.87 \\
\hline
\multicolumn{9}{c}{\textbf{GEO170K}} \\
\hline
Geometry problems & 18.67 & 26.67 & 25.33 & 14 & 38 & 40.67 & 38 & 36 \\
\hline
\multicolumn{9}{c}{\textbf{CountBench}} \\
\hline
Counting objects & 53.33 & 47.33 & 50 & 54 & 45 & 50 & 38 & 36 \\
\hline
\multicolumn{9}{c}{\textbf{Blind}} \\
\hline
Line Intersection & 35 & 39 & 32 & 37 & 33 & 33 & 31 & 40 \\
Two line intersection & 59 & 62 & 55 & 47 & 70 & 69 & 54 & 65 \\
pentagon & 12 & 13 & 2 & 13 & 6 & 9 & 4 & 2 \\
Subway & 22 & 44 & 17 & 18 & 31 & 24 & 18 & 29 \\
Rows/columns & 6 & 9 & 9 & 7 & 4 & 2 & 0 & 7 \\
Two circles & 86 & 77 & 77 & 77 & 38 & 51 & 45 & 37 \\
\hline
\textbf{Dataset Average} & 36.66 & 40.67 & 32 & 33.16 & 30.33 & 31.33 & 25.33 & 30 \\
\hline
\end{tabular}
}
\caption{Phi-3.5-Vision Model Performance Across Various Tasks}
\label{tab:florance-2-performance}
\end{table}


\end{document}